\def\cvprPaperID{3745}
\def\confName{CVPR}
\def\confYear{2023}

\def\methodname{AutoRecon}
\def\paperTitle{\methodname: Automated 3D Object Discovery and Reconstruction}

\def\authorBlock{
    Yuang Wang \quad
    Xingyi He \quad
    Sida Peng \quad
    Haotong Lin \quad
    Hujun Bao \quad
    Xiaowei Zhou$^{\dagger}$ \\[1.5mm]
    State Key Lab of CAD\&CG, Zhejiang University
}

\newif\ifreview 
\newif\ifarxiv \newcommand{\arxiv}{\arxivtrue}
\newif\ifcamera 
\newif\ifrebuttal 

\arxiv

\pdfoutput=1
\documentclass[10pt,twocolumn,letterpaper]{article}

\ifreview \usepackage[review]{cvpr} \fi
\ifarxiv \usepackage[pagenumbers]{cvpr} \fi
\ifrebuttal \usepackage[rebuttal]{cvpr} \fi
\ifcamera \usepackage{cvpr} \fi

\usepackage{graphicx}
\usepackage{amsmath}
\usepackage{amssymb}
\usepackage{booktabs}

\usepackage{times}
\usepackage{microtype}
\usepackage{epsfig}
\usepackage[table,xcdraw,dvipsname]{xcolor}
\usepackage{caption}
\usepackage{float}
\usepackage{placeins}
\usepackage{color, colortbl}
\usepackage{stfloats}
\usepackage{enumitem}
\usepackage{tabularx}
\usepackage{xstring}
\usepackage{multirow}
\usepackage{xspace}
\usepackage{url}
\usepackage{subcaption}
\usepackage[hang,flushmargin]{footmisc}
\usepackage{blindtext}
\usepackage{letltxmacro}
\usepackage{nicefrac}
\usepackage{hhline}
\usepackage{tabularx}
\usepackage{booktabs}
\usepackage{makecell}
\usepackage{scalerel}
\usepackage{suffix}

\ifcamera \usepackage[accsupp]{axessibility} \fi

\newcommand{\nbf}[1]{{\noindent \textbf{#1.}}}
\WithSuffix\newcommand\nbf*[1]{{\noindent \textbf{#1}}}

\newcommand{\supp}{supplemental material\xspace}
\ifarxiv \renewcommand{\supp}{appendix\xspace} \fi

\newcommand{\important}[1]{{\textcolor{red}{[IMPORTANT!!!]}}}

\definecolor{color1}{RGB}{249, 102, 102}
\definecolor{color2}{HTML}{579BB1}
\definecolor{color3}{RGB}{104, 185, 132}

\newcommand{\R}[1]{{%
      \textbf{%
        \ifstrequal{#1}{1}{\textcolor{color1}{R#1}}{%
          \ifstrequal{#1}{2}{\textcolor{color2}{R#1}}{%
            \ifstrequal{#1}{3}{\textcolor{color3}{R#1}}{%
              \ifstrequal{#1}{4}{\textcolor{teal}{R#1}}{%
                \textcolor{cyan}{R#1}%
              }}}}%
      }%
    }}
\WithSuffix\newcommand\R*[1]{{%
      \textbf{%
        \ifstrequal{#1}{1}{\textcolor{color1}{n4SY}}{%
          \ifstrequal{#1}{2}{\textcolor{color2}{SCgj}}{%
            \ifstrequal{#1}{3}{\textcolor{color3}{UHBX}}{%
              \ifstrequal{#1}{4}{\textcolor{teal}{R#1}}{%
                \textcolor{cyan}{R#1}%
              }}}}%
      }%
    }}
\newcounter{CQ} 
\newcounter{QR1} 
\newcounter{QR2} 
\newcounter{QR3} 
\newcommand{\printcounter}[1]{\arabic{#1}\stepcounter{#1}}
\newcommand{\Q}[2]{{%
      \noindent
      \textbf{%
        \ifstrequal{#1}{0}{{CQ.\printcounter{CQ} - #2.}}{%
          \ifstrequal{#1}{1}{{\textcolor{color1}{Q.\printcounter{QR1}} - #2.}}{%
            \ifstrequal{#1}{2}{{\textcolor{color2}{Q.\printcounter{QR2}} - #2.}}{%
              \ifstrequal{#1}{3}{{\textcolor{color3}{Q.\printcounter{QR3}} - #2.}}{%
              }}}}%
      }%
    }}

\LetLtxMacro{\oldblindtext}{\blindtext}
\LetLtxMacro{\oldBlindtext}{\Blindtext}

\newcommand\blfootnote[1]{%
  \begingroup
  \renewcommand\thefootnote{}\footnote{#1}%
  \addtocounter{footnote}{-1}%
  \endgroup
}

\usepackage{xr-hyper}

\makeatletter
\newcommand*{\addFileDependency}[1]{
  \typeout{(#1)}
  \@addtofilelist{#1}
  \IfFileExists{#1}{}{\typeout{No file #1.}}
}

\makeatother

\usepackage[pagebackref,breaklinks,colorlinks]{hyperref}
\usepackage[capitalize]{cleveref}
\crefname{section}{Sec.}{Secs.}
\crefname{table}{Table}{Tables}
\crefname{figure}{Fig.}{Figs.}

\begin{document}

\pdfstringdefDisableCommands{\let\hskip\empty}
\definecolor{citecolor}{HTML}{0071bc}

\hypersetup{
  breaklinks   = true,
  colorlinks   = true, %
  citecolor    = citecolor %
}

\twocolumn[{%
      \renewcommand\twocolumn[1][]{#1}%
      \title{\paperTitle}
      \author{\authorBlock}
      \maketitle
    }]

\begin{abstract}
A fully automated object reconstruction pipeline is crucial for digital content creation. 
While the area of 3D reconstruction has witnessed profound developments, the removal of background to obtain a clean object model still relies on different forms of manual labor, such as bounding box labeling, mask annotations, and mesh manipulations.
In this paper, we propose a novel framework named \methodname{} for the automated discovery and reconstruction of an object from multi-view images.
We demonstrate that foreground objects can be robustly located and segmented from SfM point clouds by leveraging self-supervised 2D vision transformer features.
Then, we reconstruct decomposed neural scene representations with dense supervision provided by the decomposed point clouds, resulting in accurate object reconstruction and segmentation.
Experiments on the DTU, BlendedMVS and CO3D-V2 datasets demonstrate the effectiveness and robustness of \methodname. 
The code and supplementary material are available on the project page: \url{https://zju3dv.github.io/autorecon/}.
\end{abstract}

\blfootnote{The authors are affiliated with the ZJU-SenseTime Joint Lab of 3D Vision. $^\dagger$Corresponding author: Xiaowei Zhou.}

\section{Introduction}
\label{sec:intro}

3D object reconstruction has long been investigated in computer vision.
In this work, we focus on the specific setting of reconstructing a salient foreground object from multi-view images and automatically segmenting the object from the background without any annotation, which enables scalable 3D content creation for VR/AR and may open up the possibility to generate free 2D and 3D object annotations at a large scale for supervised-learning tasks.

\begin{figure}[tp]
    \centering
    \includegraphics[width=\linewidth]{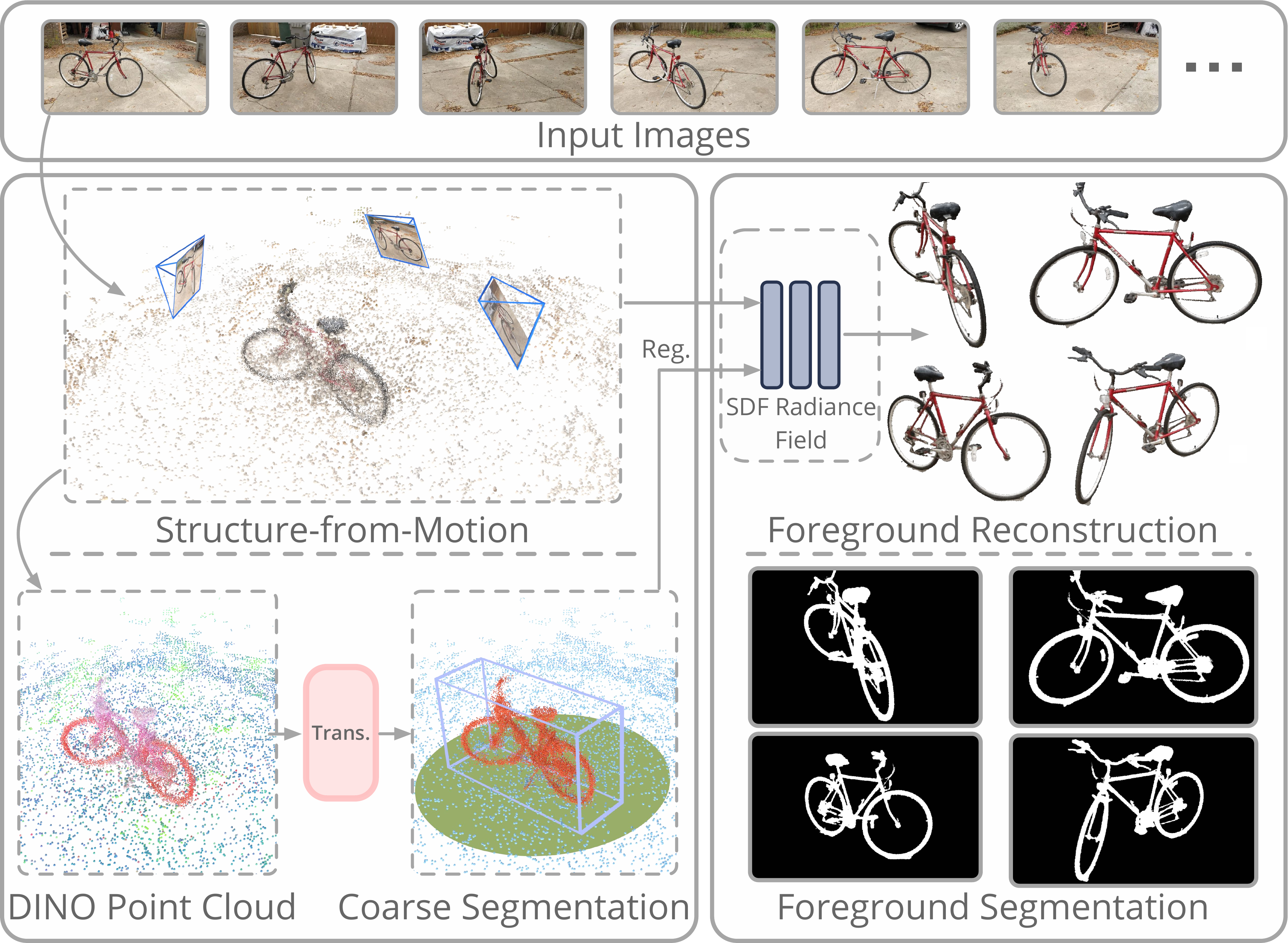}
    \caption{\textbf{Overview of our fully-automated pipeline and results.}
    Given an object-centric video, we achieve coarse decomposition by segmenting the salient foreground object from a semi-dense SfM point cloud, with pointwise-aggregated 2D DINO features~\cite{caronDINO2021}.
    Then we train a decomposed neural scene representation from multi-view images with the help of coarse decomposition results to reconstruct foreground objects and render multi-view consistent high-quality foreground masks.
    }
    \label{fig:overview}
\end{figure}

Traditional multi-view stereo \cite{furukawaInternetscaleMultiviewStereo2010,schonbergerCOLMAPMVS2016} and recent neural scene reconstruction methods \cite{wangNeuSLearningNeural2021a,yarivVolSDF2021} have attained impressive reconstruction quality.
However, these methods cannot identify objects and the reconstructed object models are typically coupled with the surrounding background.
A straightforward solution is utilizing the foreground object masks to obtain clean foreground object models.
However, accurate 2D object masks are expensive to annotate, and salient object segmentation techniques \cite{simeoniLOST2021,wangTokenCut2022,melas-kyriaziDeepSpectralMethods2022} generally produce masks with limited granularity, thus degrading the reconstruction quality, especially for objects with thin structures.
Recently, some methods \cite{zhangNeRF++2020,renNVOS2022,mirzaeiLaTeRFLabelText2022} attempt to automatically decompose objects from 3D scenes given minimal human annotations, such as 3D object bounding boxes, scribbles or pixel labels.
But the requirement of manual annotations limits the feasibility of more scalable 3D content creation.

In this paper, we propose a novel two-stage framework for the fully-automated 3D reconstruction of salient objects, as illustrated in~\cref{fig:overview}.
We first perform coarse decomposition to automatically segment the foreground SfM point cloud, and then reconstruct the foreground object geometry by learning an implicit neural scene representation under explicit supervision from the coarse decomposition.
The key idea of our coarse decomposition is to leverage the semantic features provided by a self-supervised 2D Vision Transformer (ViT) ~\cite{caronDINO2021}. Specifically, we aggregate multi-view ViT features from input images to the SfM point cloud and then segment salient foreground points with a point cloud segmentation Transformer.
To train the Transformer on large-scale unlabeled data, we devise a pseudo-ground-truth generation pipeline based on Normalized Cut \cite{shiNCut2000} and show its ability to produce accurate segmentations and 3D bounding boxes upon training.
For object reconstruction, we learn a neural scene representation within the estimated foreground bounding box from multi-view images. Our main idea is to reconstruct a decomposed scene representation with the help of explicit regularization provided by the previously decomposed point cloud.
Finally, we can extract a clean object model and obtain high-quality object masks with foreground-only rendering.

We conduct experiments on the CO3D~\cite{reizensteinCO3D2021}, BlendedMVS~\cite{yaoBlendedMVSLargescaleDataset2020}, and DTU~\cite{jensenLargeScaleMultiview2014} datasets to validate the effectiveness of the proposed pipeline. The experimental results show that our approach can automatically and robustly recover accurate 3D object models and high-quality segmentation masks from RGB videos, even with cluttered backgrounds.

In summary, we make the following contributions:
\begin{itemize}
\item We propose a fully-automated framework for reconstructing background-free object models from multi-view images without any annotation.
\item We propose a coarse-to-fine pipeline for scene decomposition by first decomposing the scene in the form of an SfM point cloud, which then guides the decomposition of a neural scene representation.
\item We propose an SfM point cloud segmentation Transformer and devise an unsupervised pseudo-ground-truth generation pipeline for its training.
\item We demonstrate the possibility of automatically creating object datasets with 3D models, 3D bounding boxes, and 2D segmentation masks.
\end{itemize}

\section{Related Work}
\label{sec:related}

\nbf{Multi-view 3D object reconstruction}
The reconstruction of 3D objects from multi-view images has long been studied with broad applications.
Aside from multi-view images, accurate object masks are needed to separate the object of interest from its surroundings and optionally provide additional geometric constraints.
Multi-view Stereo (MVS) methods~\cite{schonbergerCOLMAPMVS2016,yaoMVSNetDepthInference2018} recover a background-free reconstruction by recovering frame-wise depth maps, followed by fusing depth only within object masks.
Recently, neural reconstruction methods, built upon differentiable neural renderers and scene representations, have witnessed profound development. Surface-rendering-based methods~\cite{niemeyerDVR2020,yarivIDR2020} get rid of 3D supervision, but they still rely on object masks as substitutive geometry constraints. 
The recent volume-rendering-based reconstruction methods~\cite{yarivVolSDF2021,wangNeuSLearningNeural2021a,oechsleUNISURFUnifyingNeural2021a} allow mask-free training but still require object masks supervision to produce background-free object models.
Aside from object masks, existing methods also require manual annotation of the 3D spatial extent of the foreground object.  %
Instead, we propose a fully-automated object reconstruction pipeline without any human labeling, which further improves the usability and scalability of 3D object reconstruction.

\nbf{Decomposition of neural scene representations}
Many recent works try to decompose neural scene representations (NSR).
We categorize related works based on the annotations required.
Explicit 3D geometric primitives provide simple but effective decompositions of different entities. NeRF++~\cite{zhangNeRF++2020} separates foreground and background with a sphere. 3D bounding boxes manually annotated or predicted by category-specific models are used for decomposed modeling of static and dynamic scenes~\cite{ostNeuralSceneGraphs2021, mullerAutoRF2022, kunduPanopticNeuralFields2022}.
Multi-view segmentation masks provide dense annotations for scene decomposition.
It has been shown that semantic fields can be learned with multi-view semantic masks~\cite{kohliSRN-Seg2021,voraNeSF2021,zhiSemanticNeRF2021} for semantic scene decomposition.
Moreover, decomposed object representations can also be built from multi-view object masks~\cite{yangLearningObjectCompositionalNeural2021,wuObjectCompositionalNeuralImplicit2022a}.
To alleviate the annotation cost of multi-view segmentation, methods relying on human interactions~\cite{zhiILabelInteractiveNeural2021} and different forms of sparse human annotations are proposed, such as scribbles~\cite{renNVOS2022} and seed points~\cite{mirzaeiLaTeRFLabelText2022}. The decomposition is less stable as they rely on handcrafted non-dedicated features from various sources to distinguish the manually specified entities.
Apart from learning discrete semantic labels, DFF~\cite{kobayashiDistilledFeatureField2022} and N3F~\cite{tschernezkiN3F2022} distill 2D features into neural scene representations for query-based scene decomposition.
However, they still require manually-provided queries and their query-based nature is more suitable for local editing and impedes applications requiring global reasoning upon a scene, such as the decomposition of a salient object.
Different from existing approaches, our pipeline requires no annotation and facilitates global reasoning.

\nbf{Unsupervised object discovery}
Unsupervised object discovery (UOD) aims at the unsupervised learning of object concepts from a large-scale dataset. 
Recently, many works strive for UOD with compositional generative modeling~\cite{engelckeGENESISGenerativeScene2020,burgessMONetUnsupervisedScene2019,greffMultiObjectRepresentationLearning2020}. Slot Attention\cite{locatelloSlotAttention2020} facilitates the inference of object-centric representations directly from images. This idea is further extended to 3D-aware modeling and inference with neural radiance fields or light fields~\cite{stelznerObSuRF2021,yuUORF2022,smithCOLF2022,sajjadiOSRT2022}. 
However, these works have only been shown to work on synthetic datasets, not applicable to complex real-world situations. Our method focuses on the recovery of decomposed single-object representations and is shown to work on real-world data such as casual video captures.
Another recent trend makes use of self-supervised visual representations for unsupervised object discovery in various forms, such as object localization~\cite{simeoniLOST2021}, salient detection, and semantic segmentation~\cite{zadaianchukUnsupervisedSemanticSegmentation2022}. TokenCut~\cite{wangTokenCut2022} and DSM~\cite{melas-kyriaziDeepSpectralMethods2022} show promising results by localizing and segmenting salient objects with spectral clustering. However, their 2D nature leads to multi-view inconsistency and unstable results when applied to object-centric videos.
To overcome these limitations, we propose to perform unsupervised object discovery from videos in 3D, which facilitates coherent salient object discovery upon a global representation, instead of many isolated inferences upon local 2D observations.

\section{Preliminaries}\label{sec:pre}
In this section, we briefly review the following preliminaries: the self-supervised ViT features used to segment point clouds, the Normalized Cut algorithm employed to generate pseudo segmentation labels, and the neural surface reconstruction method NeuS utilized for object reconstruction.

\paragraph{Self-supervised ViTs.}
A Vision Transformer~\cite{dosovitskiyImageWorth16x162021} flattens an $H \times W$ sized image $\mathbf{I}$ into a sequence of  $P \times P$ sized 2D patches $\mathbf{I}_p$. Each image patch is embedded with a trainable linear projection and added with a positional embedding. A special learnable $[\mathrm{CLS}]$ token is usually prepended to the sequence of patches for modeling global and contextual information. The 1D sequence of token embeddings is fed to several Transformer encoder blocks composed of multi-head self-attention (MSA) and MLP layers:
\begin{equation} \label{TransformerEncoder}
\begin{split}
{\mathbf{z}^{\ell}}^{\prime} & = \operatorname{MSA}\left(\operatorname{LN}\left(\mathbf{z}^{\ell-1}\right)\right)+\mathbf{z}^{\ell-1}, \\
\mathbf{z}^{\ell} & = \operatorname{MLP}\left(\operatorname{LN}\left({\mathbf{z}^{\ell}}^{\prime}\right)\right)+{\mathbf{z}^{\ell}}^{\prime},
\end{split}
\end{equation}
where $\mathbf{z}^{\ell}$ is the output of the ${\ell}$-th Transformer encoder layer.

It has been shown in~\cite{caronDINO2021} that self-supervised ViT features contain explicit semantic information such as scene layout and object boundaries, which is not found in the supervised counterparts.

\begin{figure*}[tp]
    \centering
    \includegraphics[width=0.9\linewidth]{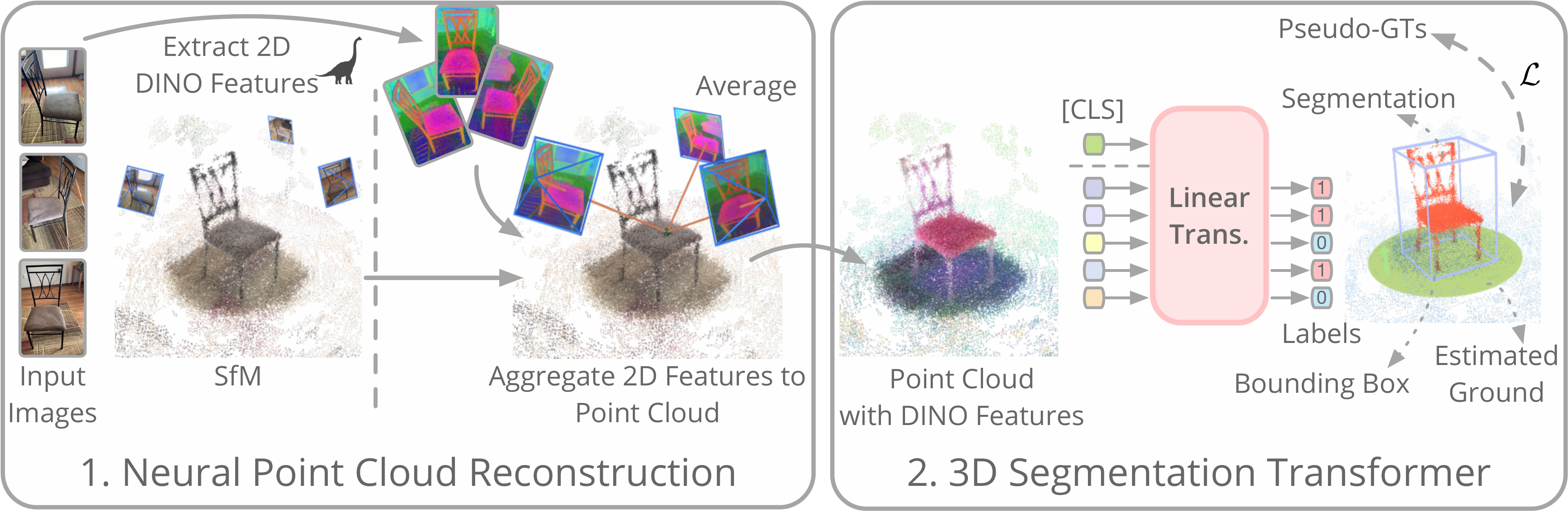}
    \caption{\textbf{Coarse Decomposition.}
    Given an object-centric image sequence, we first reconstruct the semi-dense Structure-from-Motion~(SfM) point cloud and extract pointwise features by aggregating multi-view 2D DINO features, which are semantically rich as illustrated by the PCA-projected colors.
    Then, we segment the foreground object from the SfM point cloud with a lightweight 3D Transformer, which takes pointwise features~(\protect\scalerel*{\includegraphics{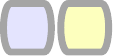}}{B}) and a global [CLS] feature~(\protect\scalerel*{\includegraphics{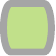}}{B}) as input and predicts pointwise labels~(\protect\scalerel*{\includegraphics{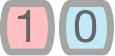}}{B}).
    Finally, the 3D bounding box of the object and an optional ground plane are estimated from the decomposed point cloud.
    }
    \label{fig:coarsedecompose}
\end{figure*}

\paragraph{Normalized cut algorithm~(NCut)~\cite{shiNCut2000}.}\label{sec:pre_ncut}
Spectral clustering is a wildly used clustering technique that originated from graph partitioning.
Given a set of data points ${x_i}$, spectral clustering builds an undirected graph $G=(V, E)$ and partitions it into two disjoint sets $A$, $B$. Each data point $x_i$ corresponds to a vertex $v_i$, and the weight $w(i, j)$ of each graph edge represents the similarity or the connectivity between two data points. Normalized Cut (NCut) is a wildly used criterion for spectral clustering, which can be efficiently minimized by solving a generalized eigenvalue problem as shown in~\cite{shiNCut2000}.

\paragraph{Neural surface reconstruction with NeuS.}
NeuS~\cite{wangNeuSLearningNeural2021a} uses the zero-level set of a signed distance function (SDF) $f: \mathbb{R}^3 \rightarrow \mathbb{R}$ to represent a surface $\mathcal{S}=\left\{\mathbf{x} \in \mathbb{R}^3 \mid f(\mathbf{x})=0\right\}$ and models appearance with a radiance field $\operatorname{c}(\mathbf{x}, \mathbf{v})$.
The SDF-based radiance field is rendered via volume rendering. Given a ray $\{ \mathbf{r}(t) = \mathbf{o} + t\mathbf{v} |~t > 0\}$, where $\mathbf{o}$ denotes the camera center and $\mathbf{v}$ is the view direction, we can render its color $\hat{C}$ by
\begin{equation} \label{eq:vol_rend}
\hat{C} = \int_{0}^{\infty} \omega (t) \operatorname{c}(\mathbf{r}(t), \mathbf{v})dt,
\end{equation}
where $\omega(t)$ is an unbiased and occlusion-aware weight function as detailed in~\cite{wangNeuSLearningNeural2021a}.

Notably, the spatial extent of the foreground object of interest needs to be manually annotated, which is scaled into a unit-sphere and represented with the SDF-based radiance field. The background region outside the sphere is represented with NeRF++~\cite{zhangNeRF++2020}. Since the object of interest can hardly be exclusively enclosed with a single sphere, the reconstructed object model includes background geometries, requiring manual post-processing for its removal.

\section{Methods}
An overview of our pipeline is illustrated in~\cref{fig:overview}.
Given an object-centric video, we aim to automatically decompose and reconstruct the salient foreground object whose high-quality 2D masks can be rendered from its reconstructed geometry.
To achieve this goal, we propose a novel coarse-to-fine pipeline that decomposes a neural scene representation with the help of point cloud decomposition.
Our coarse decomposition stage segments the foreground object from a scene-level SfM point cloud and estimates its compact 3D bounding box~(\cref{subsec:coarsedecompose}).
Then, a decomposed neural scene representation of the foreground object is recovered under explicit supervision of the coarse decomposition (\cref{subsec:finedecompose}). 

\subsection{Coarse decomposition of the salient object}
\label{subsec:coarsedecompose}

\begin{figure}[tp]
    \centering
    \includegraphics[width=0.9\linewidth]{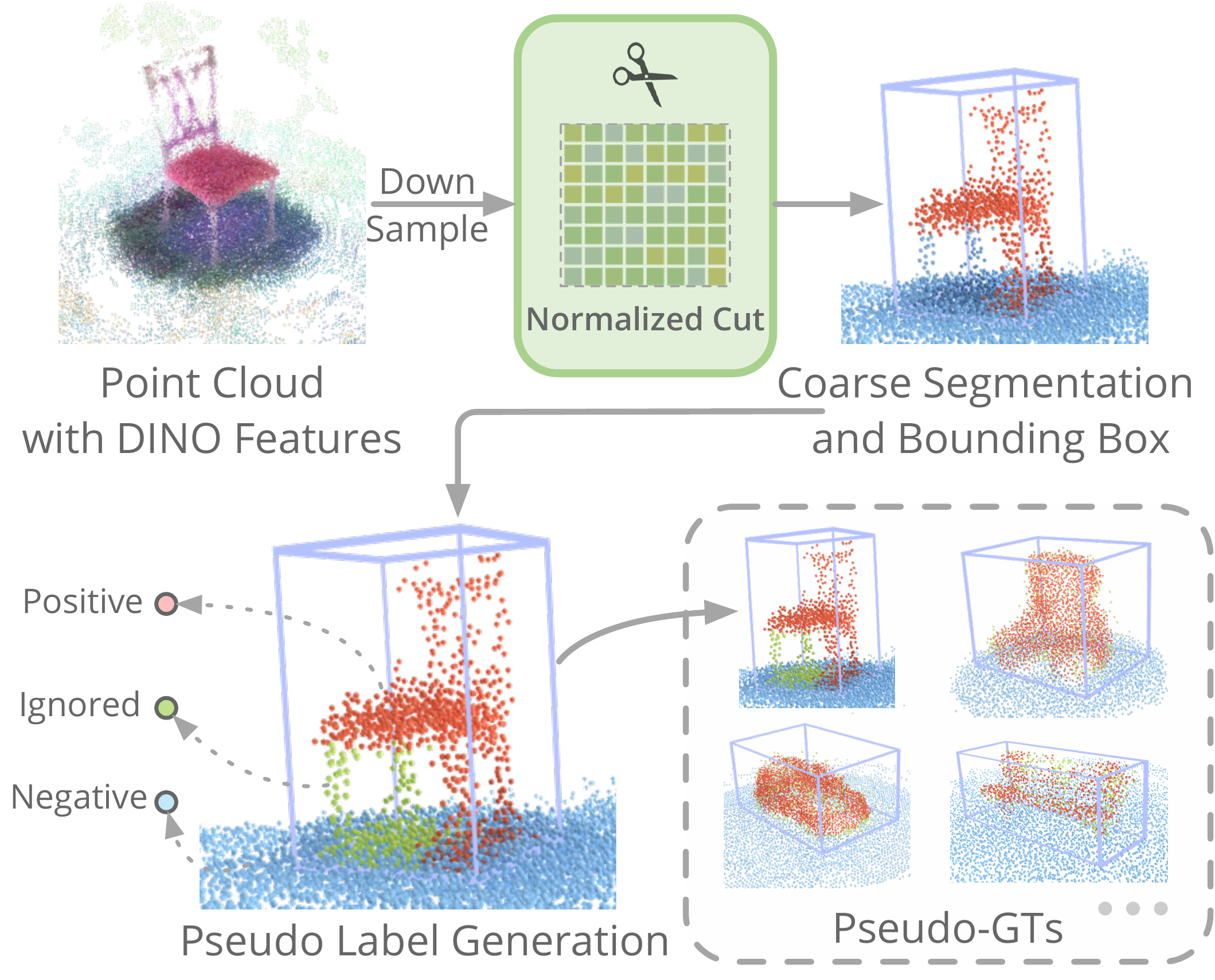}
    \caption{\textbf{Pseudo-ground-truth generation and label definition.}
    To train our point cloud segmentation Transformer with unlabeled data, we propose an unsupervised pipeline to generate pseudo labels.
    We segment the downsampled neural point cloud with Normalized Cut~\cite{shiNCut2000} (NCut) and estimate a bounding box for the foreground points.
    Taking the segmentation noise into account, we treat NCut's foreground segmentations as positive samples~(\protect\scalerel*{\includegraphics{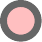}}{B}), and background points outside the bounding box as negative ones~(\protect\scalerel*{\includegraphics{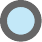}}{B}). Background segmentations located within the bounding box are regarded as segmentation noise~(\protect\scalerel*{\includegraphics{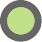}}{B}) and thus ignored in training.
    }
    \label{fig:pseudogt}
\end{figure}

To coarsely decompose the foreground object, we first reconstruct its SfM point cloud and fuse multi-view DINO~\cite{caronDINO2021} features on it.
Then, the point cloud is segmented by our lightweight 3D segmentation Transformer, upon which a 3D bounding box of the salient foreground object is generated.
Our coarse decomposition pipeline is shown in~\cref{fig:coarsedecompose}.
Since we assume that no manual annotation is available, we devise an unsupervised point cloud segmentation pipeline to generate pseudo-ground-truth segmentations, as shown in~\cref{fig:pseudogt}. Upon training, the 3D segmentation Transformer outperforms our unsupervised pipeline and can be applied to point clouds at larger scales.

\paragraph{Neural point cloud reconstruction.}
We leverage the SfM point clouds for efficient coarse decomposition since SfM is usually performed prior to the dense reconstruction for camera pose recovery.
Specifically, we use the recent semi-dense image matcher LoFTR~\cite{sunLoFTRDetectorFreeLocal2021} for SfM to reconstruct semi-dense point clouds.
It can recover the complete geometry of foreground objects, even the low-textured ones, which is discussed in~\cite{heOnePose++2022}. This capability is appealing for robustly locating an object's complete spatial extent, which is less reliable with sparse keypoints-based SfM.
To facilitate 3D segmentation, we lift self-supervised 2D ViT features to 3D.
More specifically, frame-wise DINO-ViT features are extracted and aggregated onto the semi-dense point cloud, thanks to the explicit 3D-2D correspondences retained by SfM.
We find that fusing multi-view features with a simple averaging operation is sufficient for our task.
Additionally, frame-wise features of the [CLS] token globally describing each frame are also fused as a global description of our point cloud and further used as a segmentation prototype in our proposed 3D Transformer.

\paragraph{Point cloud segmentation with Transformer.}
As shown in~\cref{fig:coarsedecompose}, the neural point cloud already contains discriminative semantic features separating foreground and background. 
Therefore, we suppose that a concise network with proper inductive bias and trained with limited supervision is enough to probe the salient object in our task.
We build an efficient point cloud Transformer with only two Transformer encoder layers and linear attentions~\cite{katharopoulosTransformersAreRNNs2020}.
The global $[\mathrm{CLS}]$ token and pointwise tokens obtained by the previously built neural point cloud are added with positional encodings and transformed by the encoder. Then, the transformed $[\mathrm{CLS}]$ token is treated as an input-dependent segmentation prototype, which is correlated with pointwise features to produce a segmentation mask. Our design takes full advantage of the global information of $[\mathrm{CLS}]$ token to reason about the salient object and its global-local relationship with other pointwise features for segmentation.
The use of pre-trained 2D ViT features alleviates the reliance on large-scale data for training Transformers.

\paragraph{Dataset generation with unsupervised segmentation.}
In order to generate training data for our 3D Transformer, we propose an unsupervised SfM segmentation pipeline, which can produce robust segmentations but is computationally more intensive.
We propose to apply NCut on the previously built neural point clouds as it facilitates scene-level global reasoning. A large-scale dataset with pseudo-ground-truth segmentations can be automatically generated with the proposed pipeline. An overview is presented in~\cref{fig:pseudogt}.

To apply the NCut algorithm on our neural point cloud for 3D segmentation,
we build a fully connected graph $G=(V, E)$ using the neural point cloud, where each graph vertex $V_i$ corresponds to a 3D point. %
We combine feature similarities and spatial affinities when modeling the edge weights $w(i, j)$ between $V_i$ and $V_j$.
Though the DINO feature is semantically rich, it is hierarchical, and the salient object inferred is sometimes ambiguous in terms of its position in the part-whole hierarchy.
We propose a grouped cosine similarity to avoid the saliency dominated by a certain object part, especially for objects with complex structures.

Formally, denote a group of multi-head attention features $\mathbf{Z}_i = \{\mathbf{z}_i^0, ..., \mathbf{z}_i^{h-1}\}$ from $h$ heads of an MSA module, we compute the grouped cosine similarity $\operatorname{S^*}$ between $\mathbf{Z}_i$ and~$\mathbf{Z}_j$:
\begin{equation} \label{eq:grouped_cos_sim}
    \operatorname{S^*} (\mathbf{Z}_i, \mathbf{Z}_j) =
    \max_{k \in \{0, ..., h-1\}}{\operatorname{S}(\mathbf{z}_i^k,\mathbf{z}_j^k)},
\end{equation}
where $\operatorname{S}$ is the cosine similarity. 
The intuition is, taking the maximum similarity between a group of multi-head features assigns two points of high similarity if they are similar in any aspect, thus reducing the chances that the saliency is only dominated by a local part of an object.
The foreground point cloud is then segmented with NCut on the graph defined above. An oriented 3D bounding box is subsequently inferred based on
plane-aligned principle component analysis of the foreground point cloud.
The pipeline is illustrated by~\cref{fig:pseudogt}.
More details about our unsupervised segmentation pipeline are provided in the \supp.

\subsection{Background-free salient object reconstruction}  %
\label{subsec:finedecompose}

\begin{figure}[tp]
    \centering
    \includegraphics[width=\linewidth]{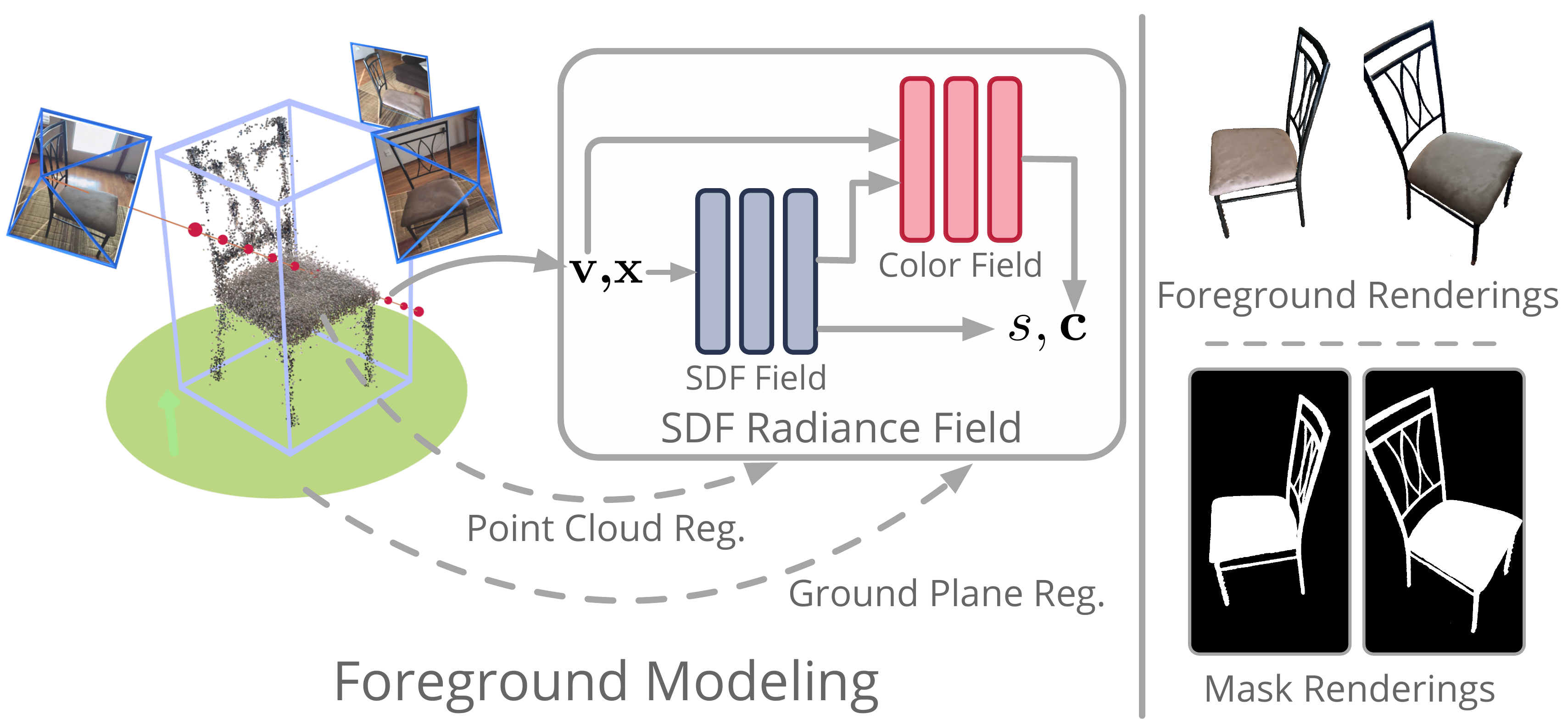}
    \caption{\textbf{Salient object reconstruction and 2D mask rendering.}
    We model the salient foreground object enclosed in the coarse bounding box with an SDF-based radiance field~\cite{wangNeuSLearningNeural2021a}.
    We use a decomposed scene representation consisting of separate fields for regions inside the bounding box, outside the bounding box, and near the ground plane.
    We regularize the optimization of the SDF-based radiance field with coarse decomposition results, i.e., the segmented foreground SfM point cloud and the estimated ground plane, for more robust foreground decomposition.
    After reconstruction, we can render high-quality multiview-consistent 2D object masks.
    }
    \label{fig:foregrounmodel}
\end{figure}

To reconstruct a background-free salient object model, we explicitly partition a scene with the coarse decomposition result and model each partition separately with neural scene representations~\cite{wangNeuSLearningNeural2021a,mildenhallNeRF2020}, which are trained upon multi-view posed images. Moreover, we incorporate multiple constraints into the optimization, which facilitates convergence and the decomposition of foreground objects from surroundings.~\cref{fig:foregrounmodel} illustrates our foreground modeling.

\paragraph{Decomposed scene modeling.}
Previous methods~\cite{yarivIDR2020,wangNeuSLearningNeural2021a} scale the foreground object into a unit-sphere with manual annotation of its spatial extent and further rely on mask culling or manual manipulation of the reconstructed mesh to remove background geometries.
Instead, we explicitly partition the scene into three parts with finer granularity without manual annotation, thanks to the estimated object bounding box in~\cref{subsec:coarsedecompose}.
More specifically, we use an SDF-based radiance field~\cite{wangNeuSLearningNeural2021a} to represent regions within the object bounding box and use a NeRF for regions outside. We use an additional tiny NeRF to model regions around the ground plane supporting the object, which can be located with the bottom plane of the object bounding box. Though there are overlaps between the inner region and the ground plane region, NeRF normally converges faster than the SDF-based radiance field and thus has an inductive bias to occupy the overlapped region.
We use a foreground-object-aware version of the contraction function with $L_{\infty}$ norm in MipNeRF-360~\cite{barronMipNeRF3602022} to model unbounded scenes. More details are provided in the \supp.

\paragraph{Explicit regularization with coarse decomposition.}
We empirically find that decomposed modeling alone cannot robustly separate a foreground object from its surroundings, especially for thin structures and regions in closed contact.
Therefore, we leverage the geometric cues in the coarse decomposition results, including the segmented foreground SfM point cloud and the estimated ground plane, to provide extra regularization for training the SDF-based radiance field.
Firstly, the unsigned distances $|f(\mathbf{x})|$ of SfM points $\mathbf{x} \in \mathbf{P}_{g}$ located on the estimated ground plane $\mathbf{P}_{g}$ are constrained to be larger than a lower bound $\theta(\mathbf{x})$:
\begin{equation}\label{eq:loss_fine_ground}
\begin{split}
    \mathcal{L}_{g} &= \frac{1}{N_g} \sum_{\mathbf{x} \in \mathbf{P}_{g}}
        \max(\theta(\mathbf{x}) - |f(\mathbf{x})|, 0), \\
    \theta(\mathbf{x}) &= \mu(\mathbf{x}) + \lambda \cdot \sigma(\mathbf{x}),
\end{split}
\end{equation}
where $\mu(\mathbf{x})$ and $\sigma(\mathbf{x})$ are means and standard deviations of unsigned distances between point $\mathbf{x}$ and its $K$ neareast neighbors.
This constraint prevents the foreground network from modeling the ground plane.

Moreover, the foreground SfM point cloud is regarded as a rough surface prior to regularize the signed distance field, similar to Geo-NeuS~\cite{fuGeoNeusGeometryConsistentNeural2022}.
This regularization can speed up convergence, alleviate the shape-radiance ambiguity and improve the reconstruction quality of thin structures.  %
Instead of directly constraining the SDF value of each SfM point to zero like in~\cite{fuGeoNeusGeometryConsistentNeural2022}, we take the noise of SfM point clouds into account.
Specifically, we model the positional uncertainty $\tau(\mathbf{x})$ of each point $\mathbf{x} \in \mathbf{P}_{fg}$ from the foreground SfM point cloud $\mathbf{P}_{fg}$ by its distances to the neighboring points similar to $\theta(\mathbf{x})$ in~\cref{eq:loss_fine_ground}.
Then we constrain the unsigned distance $|f(\mathbf{x})|$ of $\mathbf{x}$ to be smaller than $\tau(\mathbf{x})$:
\begin{equation}\label{eq:loss_fine_fg}
    \mathcal{L}_{fg} = \frac{1}{N_{fg}} \sum_{\mathbf{x} \in \mathbf{P}_{fg}}
    \operatorname{max}(|f(\mathbf{x})| - \tau(\mathbf{x}), 0).
\end{equation}

To further enhance high-quality foreground renderings with sharp boundaries, we add a beta distribution prior~\cite{lombardiNeuralVolumes2019} $\mathcal{L}_{bin}$ on the accumulated weights $O(\mathbf{r})$ of each ray $\mathbf{r} \in \mathbf{R}_{fg}$ intersecting with the object bounding box.
Finally, we use the eikonal term $\mathcal{L}_{eik}$~\cite{groppSAL2020} on sampled foreground points.
Our total loss is:
\begin{equation}\label{eq:loss_fine_all}
    \mathcal{L} = \mathcal{L}_{color} + \alpha \mathcal{L}_{eik} + \beta \mathcal{L}_{g} + \gamma \mathcal{L}_{fg} + \zeta \mathcal{L}_{bin}.
\end{equation}

\paragraph{Foreground rendering and salient object extraction.}
With the reconstructed SDF-based radiance field of the foreground object, we can easily render its multi-view consistent 2D masks and extract its mesh.
As our reconstruction models the foreground object and background with different fields, we simply compute the accumulated weights of our foreground field along each ray intersecting the object bounding box as object masks, which are binarized with a threshold of 0.5.
We use Marching Cubes~\cite{lorensenMarchingCubesHigh1987} to extract the object mesh. We can obtain a background-free object 3D model without post-processing thanks to our decomposed scene modeling.

\subsection{Implementation details}  %
\begin{figure}[tp]
    \centering
    \includegraphics[width=1.0\linewidth]{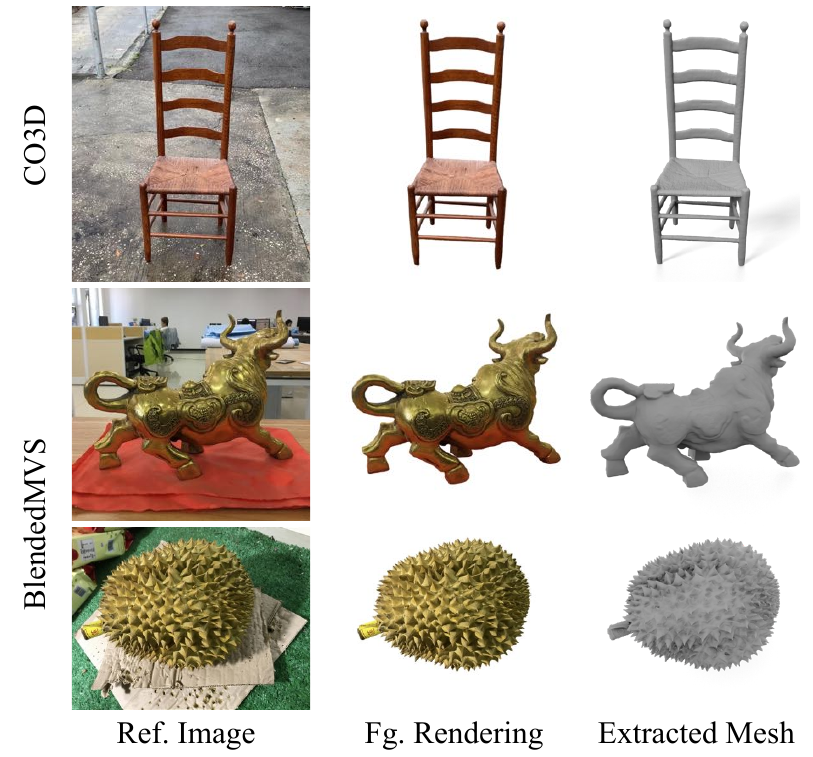}
    \caption{\textbf{Background-free salient object reconstruction results.}}
    \vspace{-0.3 cm}
    \label{fig:recon_qual}
\end{figure}

The input images used for SfM reconstruction are resized to a max area of 720,000.
We extract frame-wise 2D features from the ViT-S/8 version of DINO-ViT.
We use our data generation pipeline to process 880 object-centric videos from the CO3D-V2 dataset, which includes various categories.
All chair objects are kept as a holdout set for validation. This leads to 800 objects for training and 80 objects for validation.
We train our 3D segmentation Transformer for 20 epochs.
We use multiresolution hash encoding~\cite{mullerInstantNGP2022} and separate proposal MLPs~\cite{barronMipNeRF3602022} in all fields of our scene representation. We train our scene representation for $60k$ iterations, which takes 2 hours on a single NVIDIA V100 GPU.
All loss weights in~\cref{eq:loss_fine_all} are set to 0.1. The explicit regularization terms
are only applied during the initial $15k$ iterations with their loss weights gradually annealed to zero.

\section{Experiments}
\label{sec:experiments}
\subsection{Datasets}
We evaluate the proposed method on CO3D-V2~\cite{reizensteinCO3D2021}, BlendedMVS~\cite{yaoBlendedMVSLargescaleDataset2020} and DTU~\cite{jensenLargeScaleMultiview2014} datasets.

CO3D contains 19,000 video sequences of objects from 50 MS-COCO categories.
Many objects in CO3D contain thin structures, which are challenging for detection and segmentation from SfM point clouds.
We use the CO3D dataset to evaluate 3D salient object detection and 2D segmentation to demonstrate the capabilities of our method on challenging objects and casual captures.
To evaluate 3D detection, we manually annotate ground-truth 3D bounding boxes of 80 objects from the chair category based on the given MVS point clouds.
Moreover, we annotate detailed 2D foreground masks of $5$ objects to evaluate 2D segmentation.

BlendedMVS and DTU datasets are widely used for 3D reconstruction. We use these datasets to evaluate 3D salient object detection, reconstruction, and 2D segmentation. Since meshes provided by BlendedMVS come with backgrounds, we manually segment foreground meshes and render multi-view masks of 5 objects for evaluation. Foreground object meshes are also used for producing ground truth 3D bounding boxes.
When evaluating reconstruction results, all meshes are preprocessed with object masks.

\subsection{3D salient object detection}
In this part, we evaluate our coarse decomposition results based on the 3D bounding boxes inferred from segmented foreground point clouds. More details about generating bounding boxes can be found in \supp.

\begin{table}[t]
\centering
\resizebox{1.0\columnwidth}{!}{
\setlength\tabcolsep{2pt} %
\begin{tabular}{c||cc|cc|cc} 
\Xhline{3\arrayrulewidth}
\multirow{2}{*}{}         & \multicolumn{2}{c|}{CO3D}             & \multicolumn{2}{c|}{BlendedMVS}  & \multicolumn{2}{c}{DTU}    \\ 
\cline{2-7}
                          & AP@0.5       & AP@0.7       & AP@0.5       & AP@0.7       & AP@0.5       & AP@0.7 \\ 
\hline
\textit{TokenCut + Seg. Agg.} & 0.816 & 0.204 & 0.875 & 0.625 & 0.500  & 0.167 \\ 
\hline
\textit{Ours NCut}~(ablation)                 & 0.867  & 0.306 & \textbf{1.00} &0.75 &  \textbf{1.00}& 0.667 \\
\textit{Ours Transformer}                      & \textbf{0.908}  & \textbf{0.581}              & \textbf{1.00}         & \textbf{1.00}         & 0.833& \textbf{0.833}       \\
\Xhline{3\arrayrulewidth}
\end{tabular}
}
\vspace{-0.1 cm}
\caption{
    \textbf{Quntitative results of 3D salient object detection.}
    Our method is compared with baselines using the average precision (AP) of 3D bounding box IoU with different thresholds.
}
\vspace{-0.3 cm}
\label{tab:detection}
\end{table}

\paragraph{Baselines.}
To the best of our knowledge, there is no existing baseline that holds the same setting as our coarse decomposition pipeline, which detects 3D salient objects from SfM point clouds without manual annotation for model training.  %
Therefore, we devise two straightforward pipelines for comparison to demonstrate the effectiveness of our design.
The first baseline is TokenCut + segmentation aggregation (\textit{TokenCut + Seg. Agg.}).
We first use TokenCut~\cite{wangTokenCut2022} for 2D salient object segmentation on each image and then aggregate multi-view segmentation confidences to the SfM point cloud by averaging.
Finally, we segment the SfM point cloud with a threshold of $0.5$ to determine the salient object's 3D region.
Another baseline is our neural point cloud + NCut-based segmentation (\textit{Ours NCut}), which is used to generate pseudo-GTs for training \textit{Ours Transformer}.  %

\vspace*{-3mm}
\paragraph{Evaluation metrics.}
We use the Intersection-over-Union~(IoU) metric with thresholds of $0.5$ and $0.7$ to evaluate the bounding box accuracy. The average percision~(AP) is used for comparison.
\vspace*{-3mm}

\paragraph{Results.} %
As shown in~\cref{tab:detection}, our approach substantially achieves better salient object detection performances on all datasets, especially on the challenging CO3D~\cite{reizensteinCO3D2021} dataset.
Instead of individually segmenting 2D images as in \textit{TokenCut + Seg. Agg.}, our strategy of aggregating multi-view 2D features and performing segmentation on 3D facilitates global reasoning of the salient object and eliminates multi-view inconsistency.
The proposed \textit{Ours Transformer} also outperforms \textit{Ours NCut} baseline on most datasets and metrics although trained on pseudo-GTs generated by \textit{Ours NCut}.
We attribute this improvement to \textit{Our Transformer}'s ability to accept point clouds with a higher density as inputs, its ability to capture global dependencies, and the extra learning on the dataset generated by \textit{Ours NCut}.

\subsection{Object reconstruction and 2D segmentation}
\begin{figure}[tp]
    \centering
    \vspace{-0.2 cm}
    \includegraphics[width=1.0\linewidth]{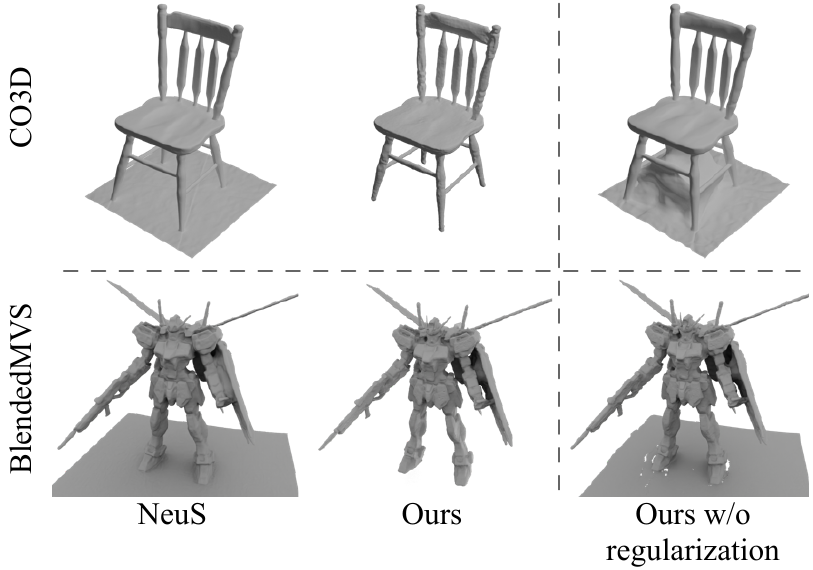}
    \vspace{-0.3 cm}
    \caption{\textbf{Qualitative results of salient object reconstruction.}
    Our method is compared with NeuS on the CO3D and BlendedMVS datasets.
    We present results with and without our explicit regularization to illustrate its effectiveness.
    }
    \label{fig:reconstruction}
\end{figure}

We evaluate the reconstructed object geometry and 2D foreground mask renderings to demonstrate the capability of our approach to reconstruct and segment complex objects.

\paragraph{Baselines.}
For 3D reconstruction, we compare our method with the neural surface reconstruction baseline NeuS~\cite{wangNeuSLearningNeural2021a}.
As for the evaluation of 2D segmentation, the proposed method is compared with following baselines in two categories: 
1) single-view image segmentation baseline TokenCut~\cite{wangTokenCut2022}, which performs 2D salient object segmentation on each image and does not consider multi-view information.
2) multi-view image segmentation baseline SemanticNeRF~\cite{zhiSemanticNeRF2021}, which fuses noisy masks with a neural field and produces high-quality masks with neural rendering. Specifically, we use the segmentations from TokenCut as inputs for SemanticNeRF and evaluate its mask renderings.
\begin{table}[tb]
    \centering
    \resizebox{1.0\columnwidth}{!}{
        \setlength\tabcolsep{2pt} %
        \begin{tabular}{c|cccccc|c}
            \Xhline{3\arrayrulewidth}
            Scan ID                           & 1              & 2              & 3              & 4              & 5              & 6              & Mean           \\
            \hline
            NeuS~(w/ annotated fg. region) & \textbf{0.390} & 0.216          & 0.245          & 0.223          & \textbf{0.345} & 0.271          & \textbf{0.282} \\
            Ours~(fully-automated)     & 0.411          & \textbf{0.200} & \textbf{0.240} & \textbf{0.218} & 0.379          & \textbf{0.264} & 0.285          \\
            \Xhline{3\arrayrulewidth}
        \end{tabular}
    }
    \vspace{-0.1 cm}
    \caption{
        \textbf{Quantitative results on the BlendedMVS dataset.} We normalize the GT mesh so that its longest side equals one.
        Results on the Chamfer $l_2$ distance are presented as percentages.
    }
    \vspace{-0.3 cm}
    \label{tab:reconstruction}
\end{table}

\paragraph{Evaluation metrics.}
We evaluate 3D reconstruction on the Chamfer $l_2$ distance.
Mask IoU and Boundary IoU~\cite{chengBoundaryIoU2021} metrics are used to evaluate 2D segmentation, with the former reflecting overall segmentation quality and the latter focusing on boundary quality.
The definitions of these metrics can be found in the \supp. 

\begin{table*}[tb]
  \centering
  \resizebox{0.9\textwidth}{!}{
    \setlength\tabcolsep{4pt} %
    \begin{tabular}{c||ccccc|c||ccccc|c||ccccc|c}
      \Xhline{3\arrayrulewidth}
                              & \multicolumn{6}{c||}{CO3D}        & \multicolumn{6}{c||}{BlendedMVS} & \multicolumn{6}{c}{DTU}                                                                                                                                                                                                                                                                \\
      \cline{1-19}
      Scan ID                 & 1                                 & 2                                & 3                       & 4              & 5              & Mean           & 1              & 2              & 3              & 4              & 5              & Mean           & 1              & 2              & 3              & 4              & 5              & Mean           \\
      \hline
                              & \multicolumn{18}{c}{Mask IoU}                                                                                                                                                                                                                                                                                                                                 \\
      \hline
      Ours                    & \textbf{0.933}                    & \textbf{0.951}                   & 0.958                   & 0.962          & 0.934          & \textbf{0.947} & 0.959          & \textbf{0.987} & 0.916          & \textbf{0.936} & \textbf{0.977} & \textbf{0.955} & \textbf{0.931} & \textbf{0.969} & \textbf{0.961} & \textbf{0.959} & 0.903          & \textbf{0.945} \\
      TokenCut~(single-view)  & 0.784                             & 0.888                            & \textbf{0.976}          & 0.975          & \textbf{0.966} & 0.918          & 0.785          & 0.904          & 0.919          & 0.855          & 0.943          & 0.881          & 0.829          & 0.921          & 0.905          & 0.955          & 0.971          & 0.916          \\
      TokenCut + SemanticNeRF & 0.825                             & 0.861                            & 0.952                   & \textbf{0.980} & 0.914          & 0.906          & \textbf{0.972} & 0.906          & \textbf{0.924} & 0.877          & 0.941          & 0.924          & 0.828          & 0.921          & 0.907          & 0.957          & \textbf{0.975} & 0.918          \\
      \hline
                              & \multicolumn{18}{c}{Boundary IoU}                                                                                                                                                                                                                                                                                                                             \\
      \hline
      Ours                    & \textbf{0.912}                    & \textbf{0.937}                   & 0.839                   & 0.771          & 0.843          & \textbf{0.860} & \textbf{0.816} & \textbf{0.914} & \textbf{0.767} & \textbf{0.896} & \textbf{0.817} & \textbf{0.842} & \textbf{0.628} & \textbf{0.842} & \textbf{0.752} & \textbf{0.707} & 0.613          & \textbf{0.877} \\
      TokenCut~(single-view)  & 0.635                             & 0.832                            & \textbf{0.877}          & \textbf{0.839} & \textbf{0.887} & 0.814          & 0.493          & 0.562          & 0.664          & 0.688          & 0.695          & 0.620          & 0.572          & 0.693          & 0.525          & 0.636          & 0.803          & 0.646          \\
      TokenCut + SemanticNeRF & 0.701                             & 0.819                            & 0.847                   & 0.822          & 0.769          & 0.792          & 0.512          & 0.578          & 0.699          & 0.730          & 0.642          & 0.632          & 0.539          & 0.633          & 0.522          & 0.661          & \textbf{0.836} & 0.638          \\
      \Xhline{3\arrayrulewidth}
    \end{tabular}
  }
  \caption{
    \textbf{Quantitative results of 2D segmentation.}
    We compare our foreground mask renderings with baselines on Mask IoU and Boundary IoU metrics on multiple datasets, including CO3D, BlendedMVS, and DTU. \methodname{} outperforms baselines on most of the scans.
  }
  \label{tab:segmentation}
  \vspace{-0.4cm}
\end{table*}

\begin{figure}[tp]
    \centering
    \includegraphics[width=1.0\linewidth]{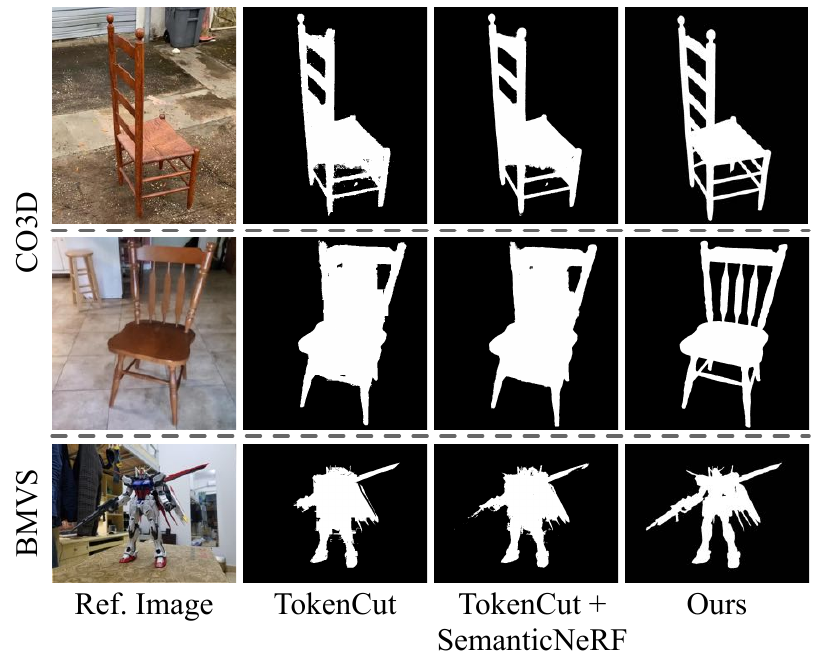}
    \vspace{-0.3 cm}
    \caption{\textbf{Qualitative results of 2D segmentation.} We show foreground segmentation on the challenging chair category in CO3D and an object in BlendedMVS with complex geometry.
    }
    \label{fig:segmentation}
\vspace{-0.3 cm}
\end{figure}

\paragraph{Results.}
For foreground object reconstruction, qualitative results are shown in~\cref{fig:recon_qual,fig:reconstruction}, and quantitative results on the BlendedMVS 
dataset are presented in~\cref{tab:reconstruction}. 
The proposed fully-automated pipeline achieves comparable or better reconstruction quality compared with NeuS, which is provided with manually-annotated object regions and requires manual post-processing for background removal. Our pipeline eliminates these tedious labors and thus demonstrates the potential to automatically create large-scale datasets.

Our method also achieves better 2D segmentation accuracy on most of the evaluated scans, as shown in~\cref{tab:segmentation} and visualized in~\cref{fig:segmentation}.
The results of the 2D salient object segmentation baseline TokenCut lack multi-view consistency and are noisy on scans with complex backgrounds. SemanticNeRF can bring improvement to the initial TokenCut segmentations on some scans.
The proposed method can handle complex objects and backgrounds and outperforms these baselines significantly on the Boundary IoU metric, which demonstrates the capability of producing high-quality segmentations.

\subsection{Ablation studies}
We conduct experiments to validate the effectiveness of our point cloud segmentation Transformer for the coarse decomposition and regularization terms for training our decomposed neural scene representation. More ablation studies are provided in the supplementary material.

\paragraph{Segmentation Transformer for coarse decomposition.}
We show the effectiveness of our 3D segmentation Transformer over our NCut-based pipeline from the higher 3D detection AP on multiple datasets, as shown in~\cref{tab:detection}.
Results show that although trained with pseudo-labels, the 3D detection accuracy improves significantly, especially on the CO3D dataset.
Moreover, \textit{Ours Transformer} runs $\sim 100\times$ times faster than \textit{Ours NCut} to segment a point cloud with $10k$ points and is applicable to large-scale point clouds.

\paragraph{Explicit regularization for training decomposed neural scene representation.}
The qualitative results in~\cref{fig:reconstruction} demonstrate the effectiveness of explicit regularization in disentangling foreground objects from their surroundings. Regularization provided by the coarse decomposition also alleviates the shape-radiance ambiguity as shown in the chair example.

\section{Conclusion}
\label{sec:conclusion}

We present a novel pipeline for fully-automated object discovery and reconstruction from multi-view images, without any human annotation.
Experiments conducted on multiple real-world datasets show the effectiveness of our method in building high-quality background-free object models.
We also demonstrate the capability of our pipeline in producing high-quality segmentation masks, which are directly applicable to 2D supervised learning. %

\paragraph{Limitations and future work.}
Problems faced by neural reconstruction methods remain in our pipeline,
like sensitivity to shadows and transient occluders and degraded results on thin-structured and non-Lambertian objects.
Storing multi-view ViT features is memory-intensive, which we expect to be alleviated by distance-preserving compression techniques.
The reconstruction quality of SfM point clouds can be further improved with refinement methods like~\cite{heOnePose++2022, lindenbergerPixPerfectSfM2021}, which can further improve the quality of surface reconstruction and potentially eliminate reconstruction ambiguities.
Our automated object reconstruction pipeline can be used to create large-scale 3D object datasets for graphics and perception tasks, such as training 2D segmentation networks and 3D generative models.

\paragraph{Acknowledgement.}
This work was supported by NSFC (No. 62172364), the ZJU-SenseTime Joint Lab of 3D Vision, and the Information Technology Center and State Key Lab of CAD\&CG, Zhejiang University.

{
  \small
  \bibliographystyle{ieee_fullname}
  \bibliography{AutoRecon}
}

\end{document}